\documentclass[journal]{IEEEtran}

\usepackage{graphicx}
\usepackage{amsmath,amssymb} 
\usepackage{color}
\usepackage[noend]{algpseudocode}
\usepackage{multirow}
\usepackage{longtable}
\usepackage[T1]{fontenc}
\usepackage{epsfig}
\usepackage{authblk}
\usepackage[symbol*]{footmisc}
\usepackage{floatrow}
\usepackage{caption}
\usepackage{hyperref}
\newfloatcommand{capbtabbox}{table}[][\FBwidth]
\ifCLASSINFOpdf
\else
\fi
\begin{document}
%
\title{Show, Attend and Translate: Unsupervised Image Translation with Self-Regularization and Attention}
%
%
%

\author{Chao Yang,~\IEEEmembership{Student Member,~IEEE,}
        Taehwan Kim,
        Ruizhe Wang,
        Hao Peng, 
        and C.-C. Jay Kuo,~\IEEEmembership{Fellow,~IEEE}
\thanks{Chao Yang and C.-C. Jay Kuo are with the Department
of Computer Science, University of Southern California, Los Angeles,
CA, 90089 USA e-mail: (see http://www.harryyang.org).}
\thanks{Taehwan Kim, Ruizhe Wang and Hao Peng are with ObEN, inc.}}
\maketitle

\begin{abstract}
Image translation between two domains is a class of problems aiming to learn mapping from an input image in the source domain to an output image in the target domain. It has been applied to numerous applications, such as data augmentation, domain adaptation, and unsupervised training. When paired training data is not accessible, image translation becomes an ill-posed problem. We constrain the problem with the assumption that the translated image needs to be perceptually similar to the original image and also appears to be drawn from the new domain, and propose a simple yet effective image translation model consisting of a single generator trained with a self-regularization term and an adversarial term. We further notice that existing image translation techniques~\cite{zhu2017unpaired,unit} are agnostic to the subjects of interest and often introduce unwanted changes or artifacts to the input. Thus we propose to add an attention module to predict an attention map to guide the image translation process. The module learns to attend to key parts of the image while keeping everything else unaltered, essentially avoiding undesired artifacts or changes. Extensive experiments and evaluations show that our model while being simpler, achieves significantly better performance than existing image translation methods.
\end{abstract}


\IEEEpeerreviewmaketitle

\section{Introduction}

\begin{figure}[!h]
\centering
\small
\setlength{\tabcolsep}{1pt}
\begin{tabular}{cc}
  \includegraphics[width=.47\textwidth]{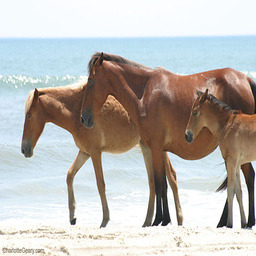}&
  \includegraphics[width=.47\textwidth]{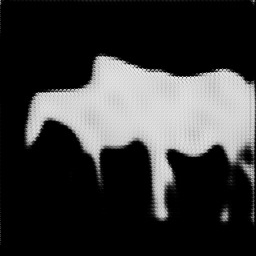}\\
  (a) Input image & (b) Predicted Attention Map \\
  \includegraphics[width=.47\textwidth]{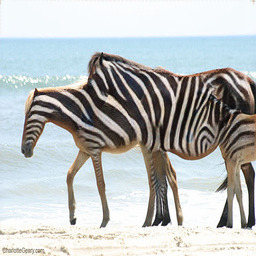}&
  \includegraphics[width=.47\textwidth]{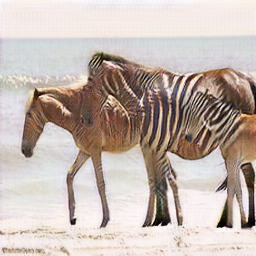} \\
(c) Final result & (d) CycleGAN~\cite{zhu2017unpaired} \\
\end{tabular}
\caption{Horse$\rightarrow$zebra image translation. Our model learns to predict an attention map (b) and translates the horse to zebra while keeping the background untouched (c). By comparison, CycleGAN~\cite{zhu2017unpaired} significantly alters the appearance of the background together with the horse (d).}.
\label{fig:teaser}
\end{figure}

Many computer vision problems can be cast as an image-to-image translation problem: the task is to map an image of one domain to a corresponding image of another domain. For example, image colorization can be considered as mapping gray-scale images to corresponding images in RGB space~\cite{zhang2016colorful}; style transfer can be viewed as translating images in one style to corresponding images with another style~\cite{gatys2016image,johnson2016perceptual,gatys2015neural}. Other tasks falling into this category include semantic segmentation~\cite{long2015fully}, super-resolution~\cite{ledig2016photo}, image manipulation~\cite{isola2016image}, etc. Another important application of image translation is related to domain adaptation and unsupervised learning: with the rise of deep learning, it is now considered crucial to have large labeled training datasets. However, labeling and annotating such large datasets are expensive and thus not scalable. An alternative is to use synthetic or simulated data for training, whose labels are trivial to acquire~\cite{zhu2017target,tzeng2015towards,rusu2016sim,richter2016playing,qiu2016unrealcv,mahendran2016researchdoom,johnson2017driving,christiano2016transfer}. Unfortunately, learning from synthetic data can be problematic and most of the time does not generalize to real-world data, due to the data distribution gap between the two domains. Furthermore, due to the deep neural networks' capability of learning small details, it is anticipated that the trained model would easily over-fits to the synthetic domain. In order to close this gap, we can either find mappings or domain-invariant representations at feature level~\cite{bousmalis2016domain,ganin2016domain,long2015learning,sun2016return,tzeng2015simultaneous,gretton2012kernel,caseiro2015beyond,ajakan2014domain,kim2017learning} or learn to translate images from one domain to another domain to create ``fake'' labeled data for training~\cite{bousmalis2017unsupervised,zhu2017unpaired,liu2017unsupervised,ledig2016photo,liu2016coupled,yoo2016pixel}. In the latter case, we usually hope to learn a mapping that preserves the labels as well as the attributes we care about.

Typically there exist two settings for image translation given two domains $X$ and $Y$. The first setting is supervised, where example image pairs $x,y$ are available. This means for the training data, for each image $x_i\in X$ there is a corresponding $y_i\in Y$, and we wish to find a translator $G: X\rightarrow Y$ such that $G(x_i)\approx y_i$. Representative translation systems in the supervised setting include domain-specific works~\cite{eigen2015predicting,hertzmann2001image,laffont2014transient,shih2013data,long2015fully,wang2016generative,xie2015holistically,zhang2016colorful} and the more general Pix2Pix~\cite{isola2016image,wang2017high}. However, paired training data comes at a premium. For example, for image stylization, obtaining paired data requires lengthy artist authoring and is extremely expensive. For other tasks like object transfiguration, the desired output is not even well defined.

Therefore, we focus on the second setting, which is unsupervised image translation. In the unsupervised setting, $X$ and $Y$ are two independent sets of images, and we do not have access to paired examples showing how an image $x_i\in X$ could be translated to an image $y_i\in Y$. Our task is then to seek an algorithm that can learn to translate between $X$ and $Y$ without desired input-output examples. The unsupervised image translation setting has greater potentials because of its simplicity and flexibility but is also much more difficult. In fact, it is a highly under-constrained and ill-posed problem, since there could be unlimited many number of mappings between $X$ and $Y$: from the probabilistic view, the challenge is to learn a joint distribution of images in different domains. As stated by the coupling theory~\cite{lindvall2002lectures}, there exists an infinite set of joint distributions that can arrive the two marginal distributions in two different domains. Therefore, additional assumptions and constraints are needed for us to exploit the structure and supervision necessary to learn the mapping. 

Existing works that address this problem assume that there are certain relationships between the two domains. For example, CycleGAN~\cite{zhu2017unpaired} assumes cycle-consistency and the existence of an inverse mapping $F$ that translates from $Y$ to $X$. It then trains two generators which are bijections and inverse to each other and uses adversarial constraint~\cite{goodfellow2014generative} to ensure the translated image appears to be drawn from the target domain and the cycle-consistency constraint to ensure the translated image can be mapped back to the original image using the inverse mapping ($F(G(x))\approx x$ and $G(F(y))\approx y$). UNIT~\cite{liu2017unsupervised}, on the other hand, assumes shared-latent space, meaning a pair of images in different domains can be mapped to some shared latent representations. The model trains two generators $G_X, G_Y$ with shared layers. Both $G_X$ and $G_Y$ maps an input to itself, while the domain translation is realized by letting $x_i$ go through part of $G_X$ and part of $G_Y$ to get $y_i$. The model is trained with an adversarial constraint on the image, a variational constraint on the latent code~\cite{kingma2013auto,rezende2014stochastic}, and another cycle-consistency constraint. 

Assuming cycle consistency ensures 1-1 mapping and avoids mode collapses~\cite{salimans2016improved}, both models generate reasonable image translation and domain adaptation results. However, there are several issues with existing approaches. First, such approaches are usually agnostic to the subjects of interest and there is little guarantee it reaches the desired output. In fact, approaches based on cycle-consistency~\cite{zhu2017unpaired,unit} could theoretically find any arbitrary 1-1 mapping that satisfies the constraints, and this renders the training unstable and the results random. This is problematic in many image translation scenarios. For example, when translating from a horse image to a zebra image, most likely we only wish to draw the particular black-white stripes on top of the horses while keeping everything else unchanged. However, what we observe is that existing approaches~\cite{zhu2017unpaired,liu2017unsupervised} do not differentiate between the horse/zebra from the scene background, and the colors and appearances of the background often significantly change during translation (Fig.~\ref{fig:teaser}). Second, most of the time we only care about one-way translation, while existing methods like CycleGAN~\cite{zhu2017unpaired} and UNIT~\cite{unit} always require training two generators of bijections. This is not only cumbersome but it is also hard to balance the effects of the two generators. Third, there is a sensitive trade-off between the faithfulness of the translated image to the input image and how similar it resembles the new domain, and it requires excessive manual tuning of the weight between the adversarial loss and the reconstruction loss to get satisfying results. 

To address the aforementioned issues, we propose a simpler yet more effective image translation model that consists of a single generator with an attention module. We first re-consider what the desired outcome of an image translation task should be: most of the time the desired output should not only resemble the target domain but also preserve certain attributes and share similar visual appearance with input. For example, in the case of horse-zebra translation~\cite{zhu2017unpaired}, the output zebra should be similar to the input horse in terms of the scene background, the location and the shape of the zebra and horse, etc. In the domain adaptation task that translates MNIST~\cite{lecun2010mnist} to USPS~\cite{denker1989neural}, we expect the output is visually similar to the input in terms of the shape and structure of the digit such that it preserves the label. Based on such observation, our model proposes to use a single generator that maps $X$ to $Y$ and is trained with a self-regularization term that enforces perceptual similarity between the output and the input, together with an adversarial term that enforces the output to appear like drawn from $Y$. Furthermore, in order to focus the translation on key components of the image and avoid introducing unnecessary changes to irrelevant parts, we propose to add an attention module that predicts a probability map as to which part of the image it needs to attend to when translating. Such probability maps, which are learned in a completely unsupervised fashion, could further facilitate segmentation or saliency detection (Fig.~\ref{fig:teaser}). Third, we propose an automatic and principled way to find the optimal weight between the self-regularization term and the adversarial term such that we do not have to manually search for the best hyper-parameter. 

Our model does not rely on cycle-consistency or shared representation assumption, and it only learns one-way mapping. Although the constraint is susceptible to oversimplify certain scenarios, we found that the model works surprisingly well. With the attention module, our model learns to detect the key objects from the background context and is able to correct artifacts and remove unwanted changes from the translated results. We apply our model on a variety of image translation and domain adaptation tasks and show that our model is not only simpler but also works better than existing methods, achieving superior qualitative and quantitative performance. To demonstrate its application in real-world tasks, we show our model can be used to improve the accuracy of face 3D morphable model~\cite{blanz1999morphable} prediction by augmenting the training data of real images with adapted synthetic images. 
\section{Related Work}

\noindent\textbf{Generative adversarial networks (GANs)} Using GAN framework~\cite{goodfellow2014generative} for generative image modeling and synthesis has gained remarkable progress recently. The basic idea of GAN training is to train a generator and a discriminator jointly such that the generator produces realistic images that confuse the discriminator. It is known that the vanilla GAN suffers from instability in training. Several techniques have been proposed to stabilize the training process and enable it to scale to higher resolution images, such as DCGAN~\cite{radford2015unsupervised}, energy-based GAN~\cite{zhao2016energy}, Wasserstein GAN (WGAN)~\cite{salimans2016improved,arjovsky2017wasserstein}, WGAN-GP~\cite{gulrajani2017improved}, BEGAN~\cite{berthelot2017began}, LSGAN~\cite{mao2016least} and the Progressive GANs~\cite{karras2017progressive}. In our work, adversarial training is the fundamental element which ensures that the output sample from the generator appears like drawn from the target domain.

\noindent\textbf{Image translation} Image translation can be seen as generating an image in target domain conditioning on an image in the source domain. Similar problems of conditional image generation include text to image translation~\cite{zhang2016stackgan,reed2016generative}, super resolution~\cite{kim2016accurate,dong2014learning,ledig2016photo}, style transfer~\cite{gatys2016image,johnson2016perceptual,li2016combining,huang2017arbitrary} etc. Based on the availability of paired training data, image translation can be either supervised (paired) or unsupervised (unpaired). Isola et al.~\cite{isola2016image} first propose a unified framework called Pix2Pix for paired image-to-image translation based on conditional GANs. Wang~\cite{wang2017high} further extends the framework to generate high-resolution images by using deeper, multi-scale networks and improved training losses. ~\cite{esser2018variational} uses variational U-Net instead of GAN for conditional image generation. UNIT~\cite{unit} and BiCycleGAN~\cite{zhu2017toward} incorporate latent code embedding into existing frameworks and enable generating randomly sampled translation results. On the other hand, when paired training data is not available, additional constraints such as cycle-consistency loss is employed~\cite{zhu2017unpaired,huang2018multimodal}. Such constraint enforces an image to map to another domain and back to itself to ensure 1-1 mapping between the two domains. However, such techniques heavily rely on ``laziness'' of the generator network and often introduce artifacts or unwanted changes to the results. Our model leverages recent advances in neural network training and employs the perceptual-based loss~\cite{johnson2016perceptual,zhang2018unreasonable} as self-regularization, such that cycle-consistency becomes unnecessary and we can also obtain more accurate translation results. 

\noindent\textbf{Attention} Recently, attention mechanism has been successfully introduced in many applications in computer vision and language processing, e.g., image captioning ~\cite{xu2015show}, text to image generation ~\cite{xu2017attngan}, visual question answering ~\cite{xu2016ask}, saliency detection ~\cite{kuen2016recurrent}, machine translation ~\cite{bahdanau2014neural} and speech recognition ~\cite{chorowski2015attention}. Attention mechanism helps models to focus on the relevant portion of the input to resolve the corresponding output without any supervision. In machine translation ~\cite{bahdanau2014neural}, it attends on relevant words in the source language to predict the current output in the target language. To generate an image from text ~\cite{xu2017attngan}, it attends on different words for the corresponding sub-region of the image. Inversely, for image captioning ~\cite{xu2015show}, image sub-regions were attended for the next generated word. In the same spirit, we propose to use an attention module to attend to the region of interest for the image translation task in an unsupervised way.

\section{Our Method}
\begin{figure*}[t]
	\centering
	\includegraphics[width=.98\linewidth]{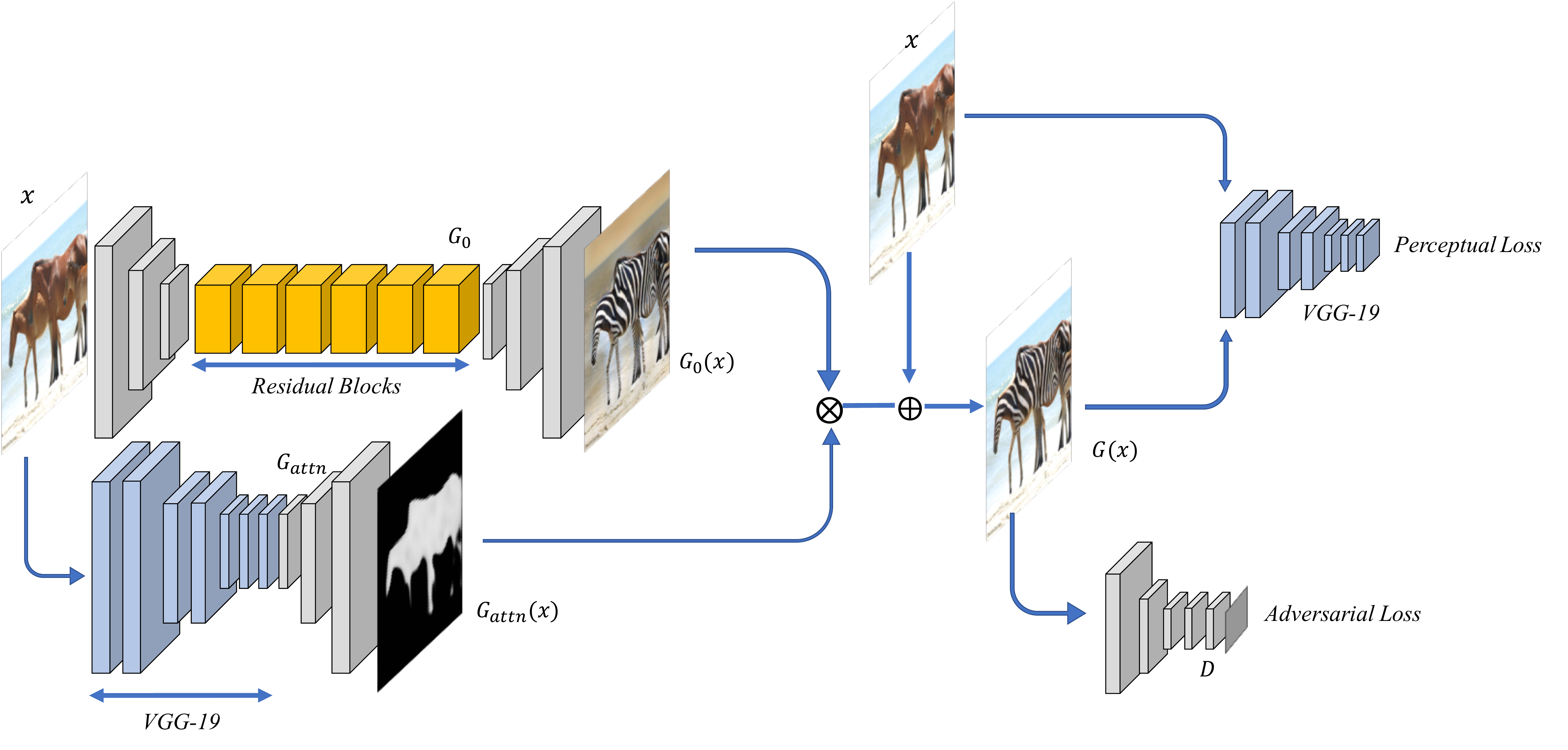}
	\caption{Model overview. Our generator $G$ consists of a vanilla generator $G_0$ and an attention branch $G_{attn}$. We train the model using self-regularization perceptual loss and adversarial loss.} 
	\vspace{-15pt}
\end{figure*}
\label{fig:framework}

We begin by explaining our model for unsupervised image translation. Let $X$ and $Y$ be two image domains, our goal is to train a generator $G_\theta: X\rightarrow Y$, where $\theta$ are the function parameters. For simplicity, we omit $\theta$ and use $G$ instead. We are given unpaired samples $x \in X$ and $y \in Y$, and the unsupervised setting assumes that $x$ and $y$ are independently drawn from the marginal distributions $P_{x\sim X}(x)$ and $P_{y\sim Y}(y)$. Let $y'=G(x)$ denote the translated image, the key requirement is that $y'$ should appear like drawn from domain $Y$, while preserving the low-level visual characteristics of $x$. The translated images $y'$ can be further used for other downstream tasks such as unsupervised learning. However, in our case, we decouple image translation from its applications.

Based on the requirements described, we propose to learn $\theta$ by minimizing the following loss:
\begin{eqnarray}
\mathcal{L}_{G} = \ell_{adv}(G(x),Y) + \lambda \ell_{reg}(x, G(x)).
\end{eqnarray}
Here $G(x)=G_{attn}(x)\otimes G_0(x) + (1-G_{attn}(x))\otimes x$, where $G_0$ is the vanilla generator and $G_{attn}$ is the attention branch. $G_0$ outputs a translated image while $G_{attn}$ predicts a probability map that is used to composite $G_0(x)$ with $x$ to get the final output. The first part of the loss, $\ell_{adv}$, is the adversarial loss on the image domain that makes sure that $G(x)$ appears like domain $Y$. The second part of the losses $\ell_{reg}$ makes sure that $G(x)$ is visually similar to $x$. In our case, $\ell_{adv}$ is given by a discriminator $D$ trained jointly with $G$, and $\ell_{reg}$ is measured with perceptual loss. We illustrate the model in Fig.~\ref{fig:framework}. 

\noindent\textbf{The model architectures:}
Our model consists of a generator $G$ and a discriminator $D$. The generator $G$ has two branches: the vanilla generator $G_0$ and the attention branch $G_{attn}$. $G_0$ translates the input $x$ as a whole to generate a similar image $G_0(x)$ in the new domain, and $G_{attn}$ predicts a probability map $G_{attn}(x)$ as the attention mask. $G_{attn}(x)$ has the same size as $x$ and each pixel is a probability value between 0-1. In the end, we composite the final image $G(x)$ by adding up $x$ and $G_0(x)$ based on the attention mask. 

$G_0$ is based on Fully Convolutional Network (FCN) and leverages properties of convolutional neural networks, such as translation invariance and parameter sharing. Similar to ~\cite{isola2016image,zhu2017unpaired}, the generator $G$ is built with three components: a down-sampling front-end to reduce the size, followed by multiple residual blocks~\cite{he2016deep}, and an up-sampling back-end to restore the original dimensions. The down-sampling front-end consists of two convolutional blocks, each with a stride of 2. The intermediate part contains nine residual blocks that keep the height/width constant, and the up-sampling back-end consists of two deconvolutional blocks, also with a stride of 2. Each convolutional layer is followed by batch normalization and ReLU activation, except for the last layer whose output is in the image space. Using down-sampling at the beginning increases the receptive field of the residual blocks and makes it easier to learn the transformation at a smaller scale. Another modification is that we adopt the dilated convolution in all residual blocks, and set the dilation factor to 2. Dilated convolutions use spaced kernels, enabling it to compute each output value with a wider view of input without increasing the number of parameters and computational burden. $G_{attn}$ consists of the initial layers of the VGG-19 network~\cite{simonyan2014very} (up to \emph{conv3\_3}), followed by two deconvolutional blocks. In the end it is a convolutional layer with sigmoid that outputs a single channel probability map. During training, the VGG-19 layers are warm-started with weights pretrained on ImageNet~\cite{russakovsky2015imagenet}.  

For the discriminator, we use a five-layer convolutional network. The first three layers have a stride of 2 followed by two convolution layers with stride 1, which effectively down-samples the networks three times. The output is a vector of real/fake predictions and each value corresponds to a patch of the image. Classifying each patch as real/fake introduces PatchGAN, and is shown to work better than the global GAN~\cite{zhu2017unpaired,isola2016image}.

\noindent\textbf{Adversarial loss:} Generative Adversarial Network~\cite{goodfellow2014generative} plays a two-player min-max game to update the network $G$ and $D$. $G$ learns to translate the image $x$ to $G(x)$ which appears as if it is from $Y$, while $D$ learns to distinguish $G(x)$ from $y$ which is the real image drawn from $Y$. The parameters of $D$ and $G$ are updated alternatively. The discriminator $D$ updates its parameters by maximizing the following objective:
\begin{eqnarray}
\mathcal{L}_D=\log (D(y))-\log(1-D (G(x))).
\label{loss_D}
\end{eqnarray}

The adversarial loss used to update the generator $G$ is defined as:
\begin{eqnarray}
\mathcal{L}_{adv}(G(x), Y)=-\log(-D(G(x))).
\label{loss_G}
\end{eqnarray}

By minimizing the loss function, the generator $G$ learns to create a translated image that fools the network $D$ into classifying the image as drawn from $Y$. 

\noindent\textbf{Self-regularization loss:}
Theoretically, adversarial training can learn a mapping $G$ that produces outputs identically distributed as the target domain $Y$. However, if the capacity is large enough, a network can map the input images to any random permutations of images in the target domain. Thus, adversarial loses alone cannot guarantee that the learned function $G$ maps the input to the desired output. To further constrain the learned mapping such that it is meaningful, we argue that $G$ should preserve visual characteristics of the input image. In other words, the output and the input need to share perceptual similarities, especially regarding the low-level features. Such features may include color, edges, shape, objects, etc. We impose this constraint with the self-regularization term, which is modeled by minimizing the distance between the translated image $y'$ and the input $x$: $\ell_{reg}=d(x, G(x))$. Here $d$ is some distance function $d$, which can be $\ell_2$, $\ell_1$, SSIM, etc. However, recent research suggests that using perceptual distance based on a pre-trained network corresponds much better to human perception of similarity comparing with traditional distance measures~\cite{zhang2018unreasonable}. In particular, we defined the perceptual loss as:
\begin{eqnarray}
\ell_{reg}(G(x),x)&&=\sum\limits_{l=1,2,3}\frac{1}{H_lW_l} \\ \nonumber
&& \sum\limits_{h,w}(\parallel w_l\circ (\hat{F}(x)^l_{hw}-\hat{F}(G(x))_{hw}^l)\parallel_2^2).
\end{eqnarray}
Here $\hat{F}$ is VGG pretrained on ImageNet used to extract the neural features; we use $l$ to represent each layer, and $H_l, W_l$ are the height and width of feature $\hat{F}^l$. We extract neural features with $\hat{F}$ across multiple layers, compute the $\ell_2$ difference at each location $h,w$ of $\hat{F}^l$ and average over the feature height and width. We then scale it with layer-wise weight $w_l$. We did extensive experiments to try different combinations of feature layers and obtained the best results by only using the first three layers of VGG and setting $w_1, w_2, w_3$ to be $1.0 / 32, 1.0 / 16, 1.0 / 8$ respectively. This conforms to the intuition that we would like to preserve the low-level traits of the input during translation. Note that this may not always be true (such as in texture transfer), but it is a hyper-parameter that could be easily adjusted based on different problem settings. We also experimented with using different pre-trained networks such as AlexNet to extract neural features as suggested by~\cite{zhang2018unreasonable} but do not observe much difference in results.

\noindent\textbf{Training scheme:}
In our experiment, we found that training the attention branch and the vanilla generator branch is difficult as it is hard to balance the learned translation and mask. In our practice, we train the two branches separately. First, we train the vanilla generator $G_0$ without the attention branch. After it converges, we train the attention branch $G_{attn}$ while keeping the trained generator $G_0$ fixed. In the end, we jointly fine-tune them with a smaller learning rate. 

\noindent\textbf{Adaptive weight induction:}
Like other image translation methods, the resemblance to the new domain and faithfulness to the original image is a trade-off. In our model, it is determined by the weight $\lambda$ of the self-regularization term relative to the image adversarial term. If $\lambda$ is too large, the translated image will be close to the input but does not look like the new domain. If $\lambda$ is too small, the translated image would fail to pertain the visual traits of the input. Previous approaches usually decide the weight heuristically. Here we propose an adaptive scheme to search for the best $\lambda$: we start by setting $\lambda=0$, which means we only use the adversarial constraint to train the generator. Then we gradually increase $\lambda$. This would lead to the increase of the adversarial loss as the output would shift away from $Y$ to $X$, which makes it easier for $D$ to classify. We stop increasing $\lambda$ when the adversarial loss reaches above some threshold $\ell_{adv}^t$. We then keep $\lambda$ constant and continue to train the network until converging. Using the adaptive weight induction scheme avoids manual tuning of $\lambda$ for each specific task and gives results that are both similar to the input $x$ and the new domain $Y$. Note that we repeat such process both when training $G_0$ and $G_{attn}$.

\noindent\textbf{Analysis:}
Our model is related to CycleGAN in that if we assume 1-1 mapping, we can define an inverse mapping $F:Y\rightarrow X$ such that $F(G(x))=x$. This satisfies the constraints of CycleGAN in that the cycle-consistency loss is zero. This shows that our learned mapping belongs to the set of possible mappings given by CycleGAN. On the other hand, although CycleGAN tends to learn the mapping such that the visual distance between $y'$ and $x$ is small possibly due to cycle-consistency constraint, it does not guarantee to minimize the perceptual distance between $G(x)$ and $x$. Comparing with UNIT, if we add another constraint that $G(y)=y$, then it is a special case of the UNIT model where all layers of the two generators are shared which leads to a single generator $G$. In this case, the cycle-consistency constraint is implicit as $G(G(x))=G(x)$ and $\min d(x,G(x))=\min d(x,G(G(x)))$. However, we observe that adding the additional self-mapping constraint for domain $Y$ does not improve the results. 

Even though our approach assumes the perceptual distance between $x$ and its corresponding $y\in Y$ is small, our approach generalizes well to tasks where the input and output domains are significantly different, such as translation of photo to map, day to night, etc., as long as our assumption generally holds. For example, in the case of photo to map, the park (photo) is labeled as green (map) and the water (photo) is labeled as blue (map), which provides certain low-level similarities. Experiments show that even without the attention branch, our model produces results consistently similar or better than other methods. This indicates that the cycle-consistency assumption may not be necessary for image translation. Note that our approach is a meta-algorithm, and we could potentially improve the results by using new/more advanced components. For example, the generator and discriminator could be easily replaced with the latest GAN architectures such as LSGAN~\cite{mao2017least}, WGAN-GP~\cite{gulrajani2017improved}, or adding spectral normalization~\cite{miyato2018spectral}. We may also improve the results by employing a more specific self-regularizaton term that is fine-tuned on the datasets we work on. 

\renewcommand{\thefootnote}{\arabic{footnote}}
\section{Results}

\begin{figure*}[h]
\centering
\small
\setlength\tabcolsep{1pt}
\begin{tabular}{cccccc}
  \includegraphics[width=.16\textwidth]{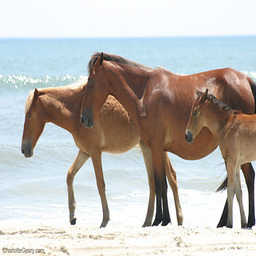}&
  \includegraphics[width=.16\textwidth]{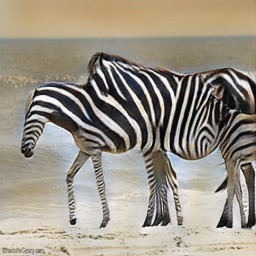}&
  \includegraphics[width=.16\textwidth]{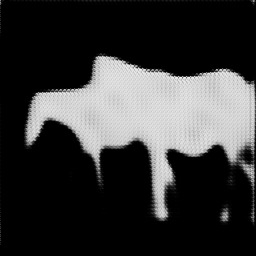}&
  \includegraphics[width=.16\textwidth]{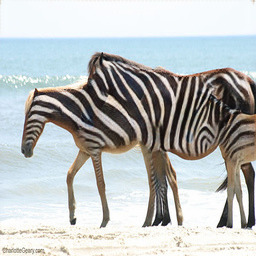}&
  \includegraphics[width=.16\textwidth]{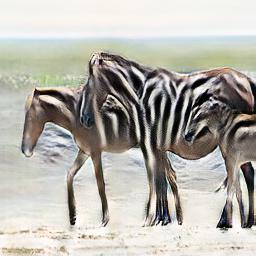}&
  \includegraphics[width=.16\textwidth]{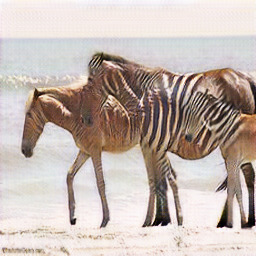} \\
  \includegraphics[width=.16\textwidth]{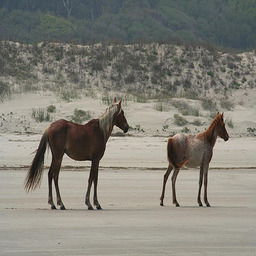}&
  \includegraphics[width=.16\textwidth]{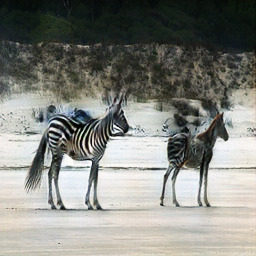}&
  \includegraphics[width=.16\textwidth]{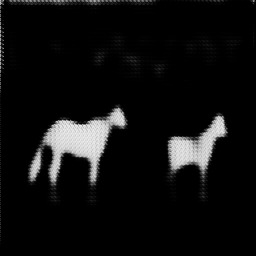}&
  \includegraphics[width=.16\textwidth]{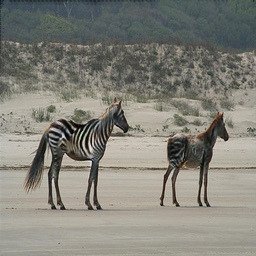}&
  \includegraphics[width=.16\textwidth]{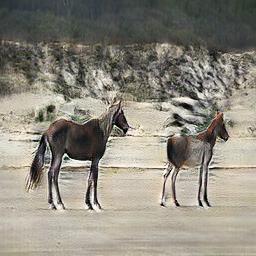}&
  \includegraphics[width=.16\textwidth]{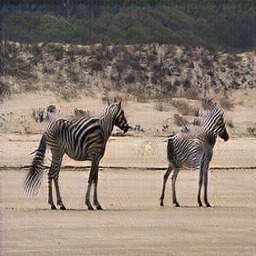} \\
  \includegraphics[width=.16\textwidth]{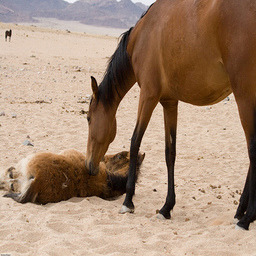}&
  \includegraphics[width=.16\textwidth]{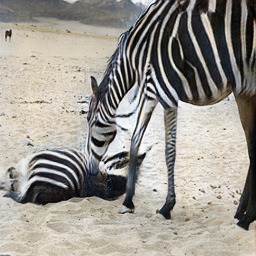}&
  \includegraphics[width=.16\textwidth]{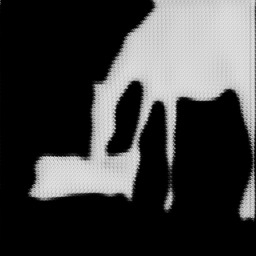}&
  \includegraphics[width=.16\textwidth]{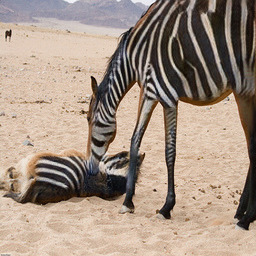}&
  \includegraphics[width=.16\textwidth]{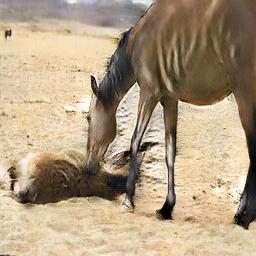}&
  \includegraphics[width=.16\textwidth]{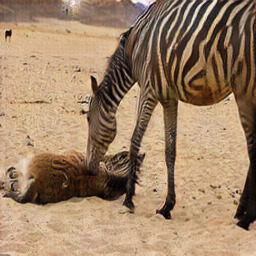} \\
  \includegraphics[width=.16\textwidth]{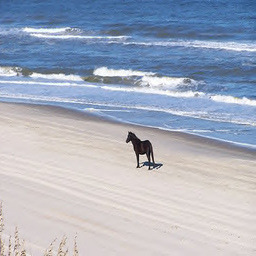}&
  \includegraphics[width=.16\textwidth]{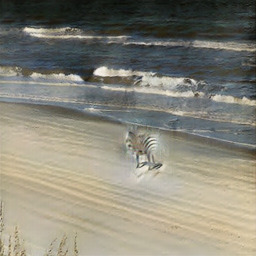}&
  \includegraphics[width=.16\textwidth]{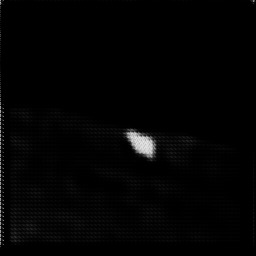}&
  \includegraphics[width=.16\textwidth]{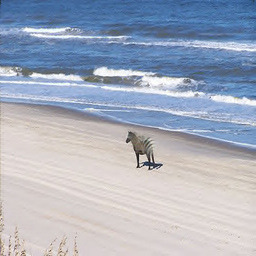}&
  \includegraphics[width=.16\textwidth]{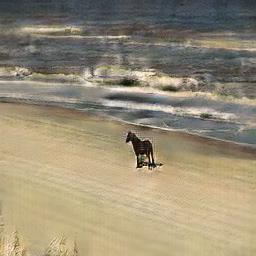}&
  \includegraphics[width=.16\textwidth]{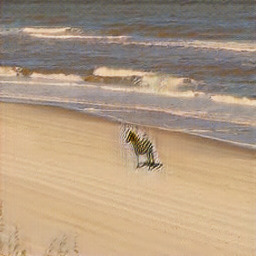} \\
  \includegraphics[width=.16\textwidth]{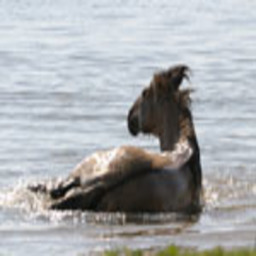}&
  \includegraphics[width=.16\textwidth]{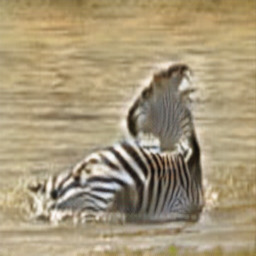}&
  \includegraphics[width=.16\textwidth]{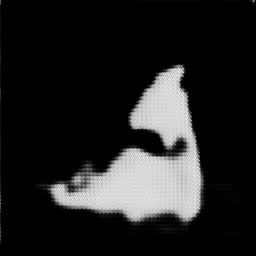}&
  \includegraphics[width=.16\textwidth]{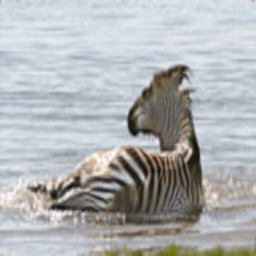}&
  \includegraphics[width=.16\textwidth]{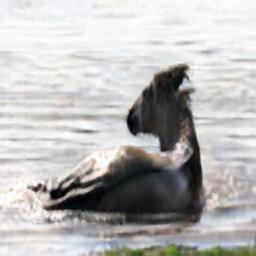}&
  \includegraphics[width=.16\textwidth]{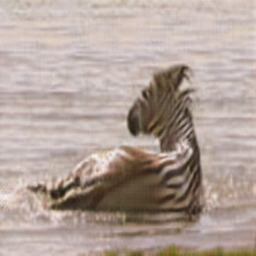} \\
 
  (a) Input  & (b) Initial trans  & (c) Attention map & (d) Final result & (e) UNIT~\cite{unit} & (f) CycleGAN~\cite{zhu2017unpaired}\\
\end{tabular}
\caption{Image translation results of horse to zebra~\cite{isola2016image} and comparison with UNIT and CycleGAN.}
\label{fig:compare}
\end{figure*}

We tested our model on a variety of datasets and tasks. In the following, we show the qualitative results of image translation, as well as quantitative results in several domain adaptation settings. In our experiments, all images are resized to 256x256. We use Adam solver~\cite{kingma2014adam} to update the model weights during training. In order to reduce model oscillation, we update the discriminators using a history of generated images rather than the ones produced by the latest generative models~\cite{shrivastava2017learning}: we keep an image buffer that stores the 50 previously generated images. All networks were trained from scratch with a learning rate of 0.0002. Starting from 5k iteration, we linearly decay the learning rate over the remaining 5k iterations. Most of our training takes about 1 day to converge on a single Titan X GPU. 

\subsection{Qualitative Results}

\begin{figure*}[!h]
\centering
\small
\setlength\tabcolsep{1pt}
\begin{tabular}{cccccccc}
  \includegraphics[width=.12\textwidth]{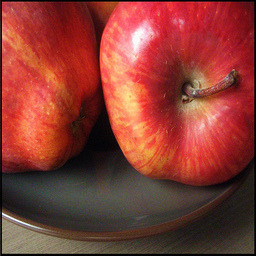}&
  \includegraphics[width=.12\textwidth]{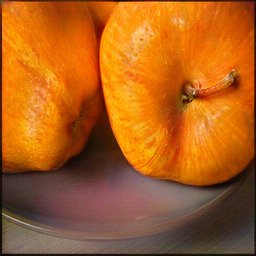}&
  \includegraphics[width=.12\textwidth]{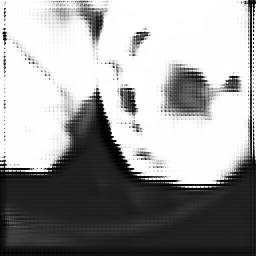}&
  \includegraphics[width=.12\textwidth]{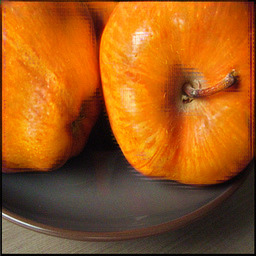}&
   \includegraphics[width=.12\textwidth]{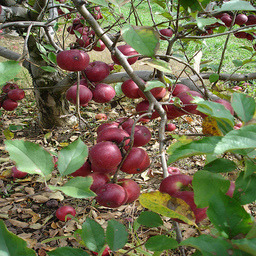}&
  \includegraphics[width=.12\textwidth]{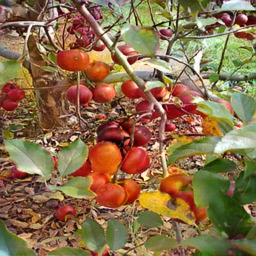}&
  \includegraphics[width=.12\textwidth]{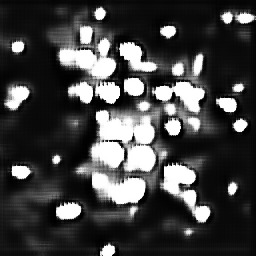}&
  \includegraphics[width=.12\textwidth]{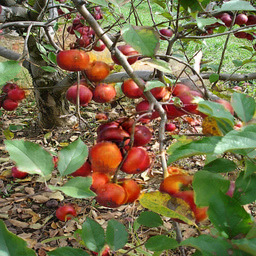} \\
  \includegraphics[width=.12\textwidth]{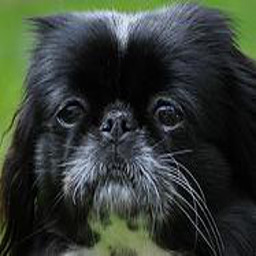}&
  \includegraphics[width=.12\textwidth]{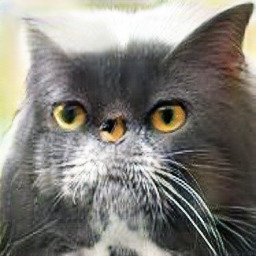}&
  \includegraphics[width=.12\textwidth]{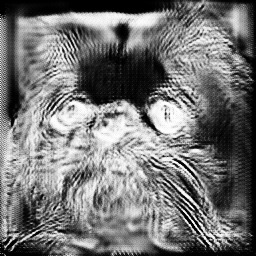}&
  \includegraphics[width=.12\textwidth]{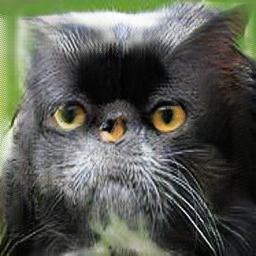}&
  \includegraphics[width=.12\textwidth]{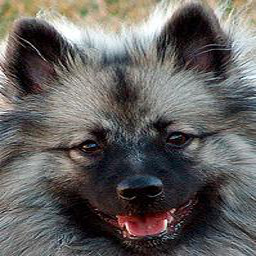}&
  \includegraphics[width=.12\textwidth]{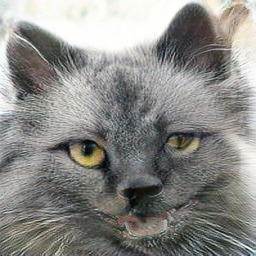}&
  \includegraphics[width=.12\textwidth]{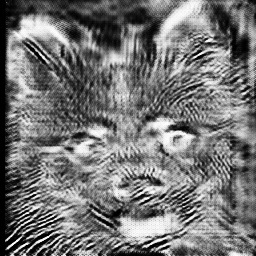}&
  \includegraphics[width=.12\textwidth]{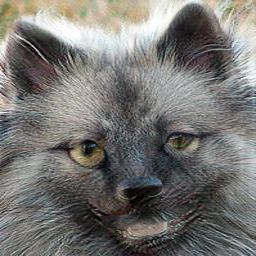} \\
  \includegraphics[width=.12\textwidth]{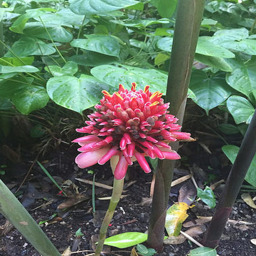}&
  \includegraphics[width=.12\textwidth]{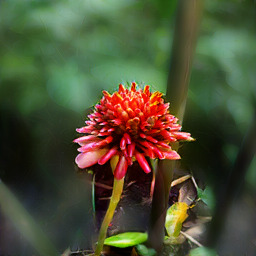}&
  \includegraphics[width=.12\textwidth]{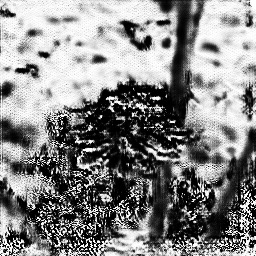}&
  \includegraphics[width=.12\textwidth]{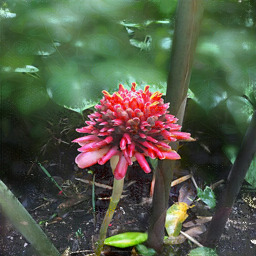}&
  \includegraphics[width=.12\textwidth]{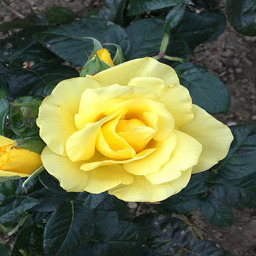}&
  \includegraphics[width=.12\textwidth]{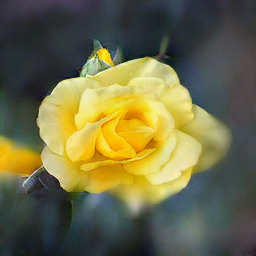}&
  \includegraphics[width=.12\textwidth]{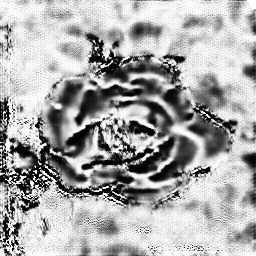}&
  \includegraphics[width=.12\textwidth]{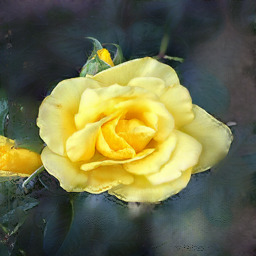} \\
  \includegraphics[width=.12\textwidth]{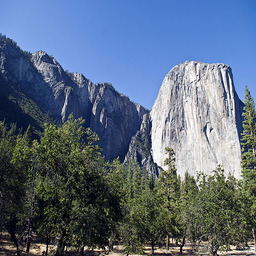}&
  \includegraphics[width=.12\textwidth]{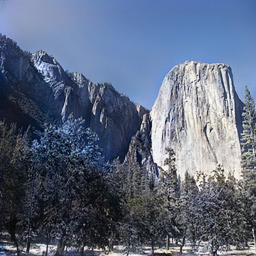}&
  \includegraphics[width=.12\textwidth]{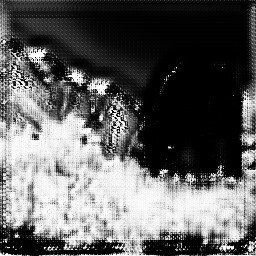}&
  \includegraphics[width=.12\textwidth]{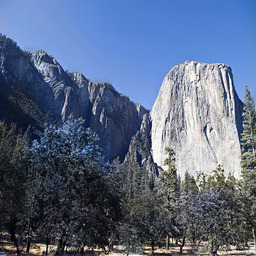}&
  \includegraphics[width=.12\textwidth]{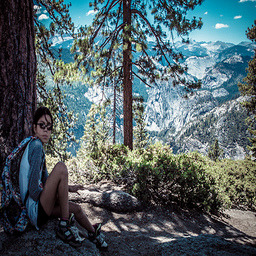}&
  \includegraphics[width=.12\textwidth]{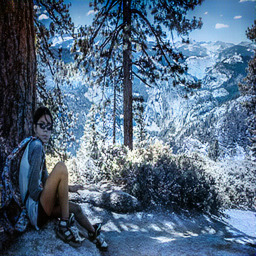}&
  \includegraphics[width=.12\textwidth]{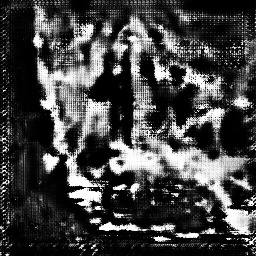}&
  \includegraphics[width=.12\textwidth]{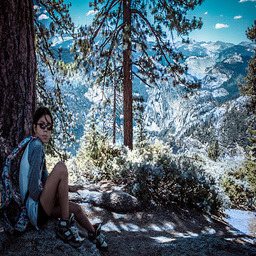} \\
  (a) Input & (b) Initial & (c) Attention & (d) Final & (e) Input & (f) Initial & (g) Attention & (h) Final \\
\end{tabular}
\caption{Image translation results on more datasets. From top to bottom: apple to orange~\cite{isola2016image}, dog to cat~\cite{parkhi12a}, photo to DSLR~\cite{isola2016image}, Yosemite summer to winter~\cite{isola2016image}.}
\label{fig:more}
\end{figure*}

\begin{figure*}[!h]
\centering
\small
\setlength\tabcolsep{1pt}
\begin{tabular}{cccccc}
\includegraphics[width=.16\textwidth]{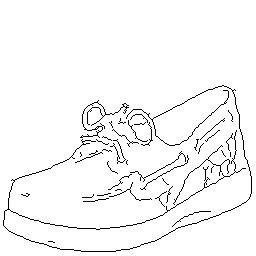}&
\includegraphics[width=.16\textwidth]{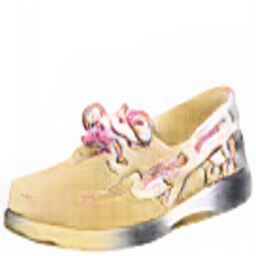}&
\includegraphics[width=.16\textwidth]{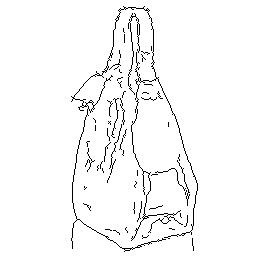}&
\includegraphics[width=.16\textwidth]{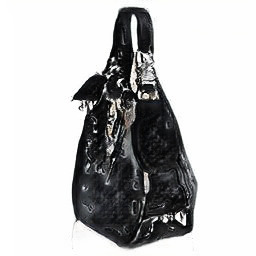}&
\includegraphics[width=.16\textwidth]{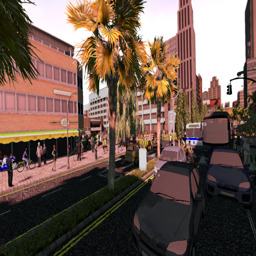}&
\includegraphics[width=.16\textwidth]{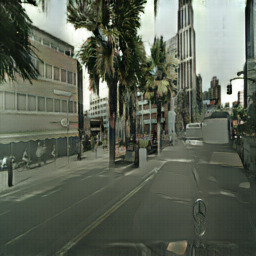}\\
\end{tabular}
\caption{More image translation results. From left to right: edges to shoes~\cite{isola2016image}; edges to handbags~\cite{isola2016image}; SYNTHIA to cityscape~\cite{ros2016synthia,cordts2015cityscapes}. Given the source and target domains are globally different, the initial translation and final result are similar with the attention maps focusing on the entire images.}
\label{fig:compare2}
\end{figure*}

Fig.~\ref{fig:compare} shows visual results of image translation of horse to zebra. For each image, we show the initial translation $G_0(x)$, the attention map $G_{attn}(x)$ and the final result $G(x)$ composited using $G_0(x)$ and $x$ based on $G_{attn}(x)$. We also compare the results with CycleGAN~\cite{zhu2017unpaired} and UNIT~\cite{unit}, and all models are trained using the same number of iterations. For the baseline implementation, we use the original authors' implementations. We can see from the examples that without the attention branch, our simple translation model $G_0$ already gives results similar or better than~\cite{zhu2017unpaired,unit}. However, all these results suffer from perturbations of background color/texture and artifacts near the region of interest. With the predicted attention map which learns to segment the horses, our final results have much higher visual quality, with the background keeping untouched and artifacts near the ROI removed (row 2, 4). Complete results of horse-zebra translations and comparisons are available online \footnote{\url{http://www.harryyang.org/img\_trans}}.

Fig.~\ref{fig:more} shows more results on a variety of datasets. We can see that for all these tasks, our model can learn the region of interest and generate compositions that are not only more faithful to the input, but also have fewer artifacts. For example, in dog to cat translation, we notice most attention maps have large values around the eyes, indicating the eyes are key ROI to differentiate cats from dogs. In the examples of photo to DSLR, the ROI should be the background that we wish to defocus, while the initial translation changes the color of the foreground flower in the photo. The final result, on the other hand, learns to keep the color of the foreground flower. In the second example of summer to winter translation, we notice the initial result incorrectly changes color of the person. With the guidance of attention map, the final result removes such artifacts.  

In a few scenarios, the attention map is less useful as the image does not explicitly contain region of interest and should be translated everywhere. In this case, the composited results largely rely on the initial prediction given by $G_0$. This is true for tasks like edges to shoes/handbags, SYNTHIA to cityscape (Fig.~\ref{fig:compare2}) and photo to map (Fig.~\ref{fig:mapf}). Although many of these tasks have very different source and target domains, our method is general and can be applied to get satisfying results. 

\begin{figure}[!h]
\centering
\small
\setlength\tabcolsep{1.5pt}
\begin{tabular}{cccc}
 \includegraphics[width=.23\textwidth]{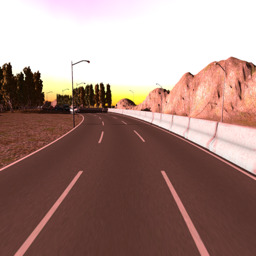}&
  \includegraphics[width=.23\textwidth]{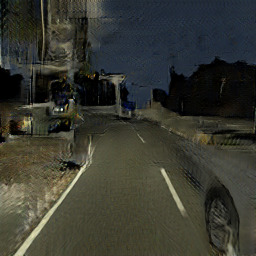}&
  \includegraphics[width=.23\textwidth]{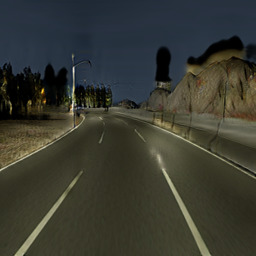}&
  \includegraphics[width=.23\textwidth]{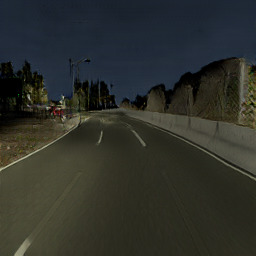} \\
  \includegraphics[width=.23\textwidth]{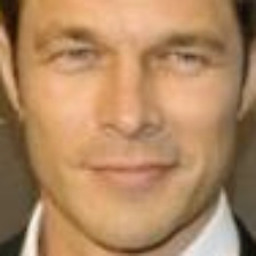}&
  \includegraphics[width=.23\textwidth]{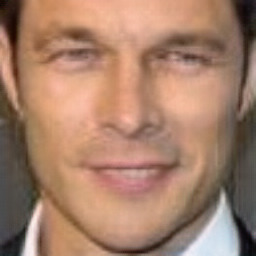}&
  \includegraphics[width=.23\textwidth]{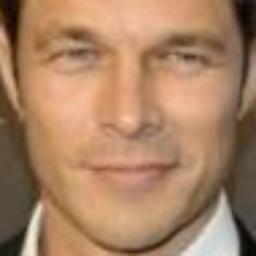}&
  \includegraphics[width=.23\textwidth]{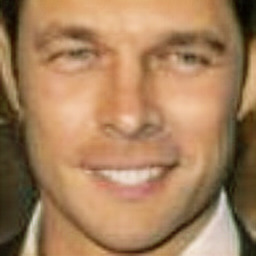} \\
    \includegraphics[width=.24\textwidth]{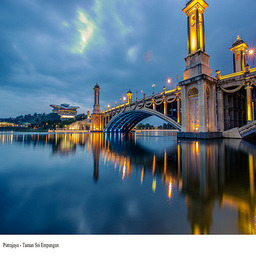}&
  \includegraphics[width=.24\textwidth]{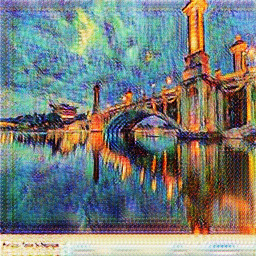}&
  \includegraphics[width=.24\textwidth]{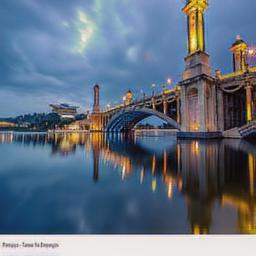}&
  \includegraphics[width=.24\textwidth]{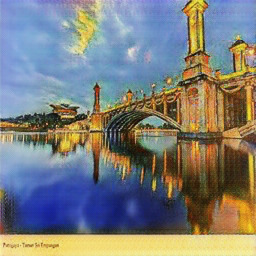} \\
  (a) Input & (b) CycleGAN  & (c) {UNIT} & (d) Ours w/o attn  \\
\end{tabular}
\caption{Comparing our results w/o attention with baselines. From top to bottom: dawn to night (SYNTHIA~\cite{ros2016synthia}), non-smile to smile (CelebA~\cite{liu2015deep}) and photos to Van Gogh~\cite{isola2016image}.}
\label{fig:no_attention}
\end{figure}

To better demonstrate the effectiveness of our simple model, Fig.~\ref{fig:no_attention} shows several results before training with the attention branch and compares with baseline. We can see that even without the attention branch, our model generates better qualitative results comparing with CycleGAN and UNIT (more samples of photo to Van Gogh is available online \footnote{\url{ http://www.harryyang.org/img_trans/vangogh}}).

\begin{figure}[!h]
\centering
\small
\setlength\tabcolsep{1pt}
\begin{tabular}{cccc}
\includegraphics[width=.24\textwidth]{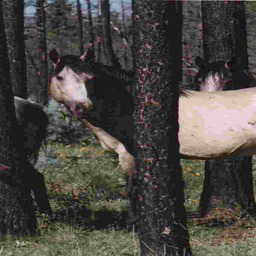}&
\includegraphics[width=.24\textwidth]{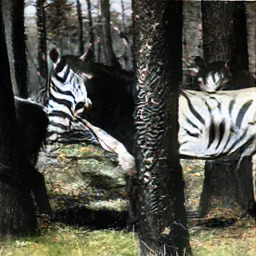}&
\includegraphics[width=.24\textwidth]{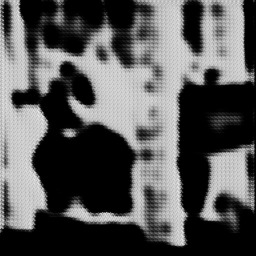}&
\includegraphics[width=.24\textwidth]{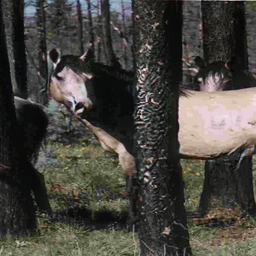} \\
\end{tabular}
\caption{Failure case of the attention map: it did not detect the ROI correctly and removed the zebra stripes.}
\label{fig:error}
\end{figure}

\begin{figure}[!h]
\centering
\small
\setlength\tabcolsep{1pt}
\begin{tabular}{cccc}
\includegraphics[width=.24\textwidth]{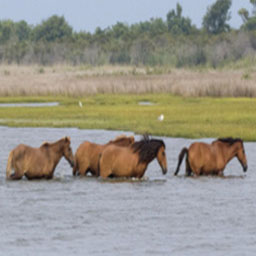}&
\includegraphics[width=.24\textwidth]{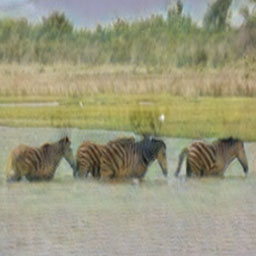}&
\includegraphics[width=.24\textwidth]{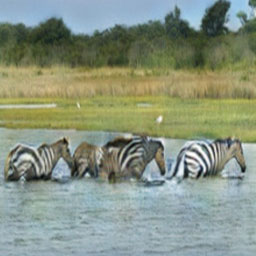}&
\includegraphics[width=.24\textwidth]{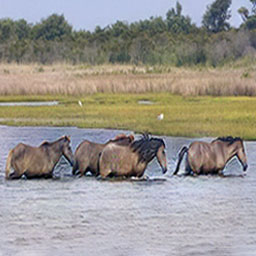} \\
(a) & (b) & (c) & (d) \\
\end{tabular}
\caption{Effects of using different layers as feature extractors. From left to right: input (a), using the first two layers of VGG (b), using the last two layers of VGG (c) and using the first three layers of VGG (d).}
\label{fig:ablation}
\end{figure}

\noindent\textbf{User study:} To more rigorously evaluate the performance, we perform a user study to compare the results. The procedure is as following: we asked for feedbacks from 22 users (all are graduate students and researchers). Each user is given 30 sets of images to compare. Each set has 5 images, which are the input, initial result (w/o attention), final result (with attention), CycleGAN results and UNIT results. In total there are 300 different image sets randomly selected from horse to zebra and photo to Van Gogh translation tasks. The images in each set are in random order. The user is then asked to rank the four results from highest visual quality to lowest. The user is fully informed about the task and is aware of the goal as to translate the input image into a new domain while avoiding unnecessary changes. 

Table~\ref{table:user_study} shows the user-study results. We listed results of: CycleGAN vs ours initial/final; UNIT vs ours initial/final; and ours initial vs ours final. We can see that our results, even without applying the attention branch (\emph{ours initial}), achieve higher ratings than CycleGAN or UNIT. The attention branch also significantly improves the results (\emph{Ours final}). In terms of directly evaluating the effects of attention branch, ours final is overwhelmingly better than ours initial based on user rankings (Table~\ref{table:user_study} row 5). We further examined the few cases where the attention results receive lower scores, and we found that the reason is due to incorrect attention maps (Fig.~\ref{fig:error}). 

\begin{table}[!h]
\begin{center}
\resizebox{1\textwidth}{!}{%
{
\setlength\tabcolsep{1.5pt}
  \begin{tabular}{ l l  c c c }
    \hline
    \textbf{Method 1} & \textbf{Method 2} & \textbf{1 better} & \textbf{About same} & \textbf{2 better} \\ \hline
    \multirow{2}{*}{Ours initial} & CycleGAN & {43.6\%} & {30.0\%} & {26.4\%} \\ \cline{2-5}
    & UNIT & {77.4\%} & {17.5\%} & {5.1\%}\\ \hline
    \multirow{3}{*}{Ours final} & CycleGAN & {63.0\%} & {21.9\%} & {15.1\%}\\ \cline{2-5}
    & UNIT & {83.8\%} & {14.4\%} & {1.8\%} \\ \cline{2-5}
    & Ours initial & {74.2\%} & {18.5\%} & {7.3\%} \\ \hline
  \end{tabular}}
  }
  \end{center}
  \caption{User study results.}
  \label{table:user_study}
\end{table}

\begin{figure*}[!h]
\begin{floatrow}

\capbtabbox{%
\setlength\tabcolsep{1pt}
\begin{tabular}{ccccc}
\includegraphics[width=.14\textwidth]{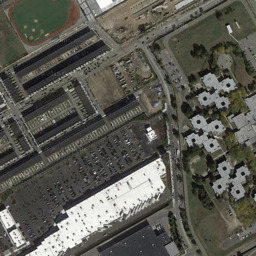}&
\includegraphics[width=.14\textwidth]{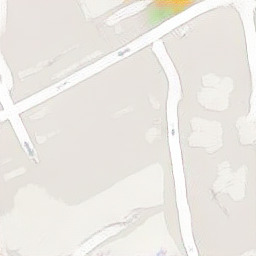}&
\includegraphics[width=.14\textwidth]{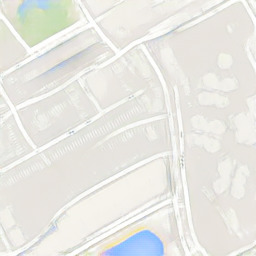}&
\includegraphics[width=.14\textwidth]{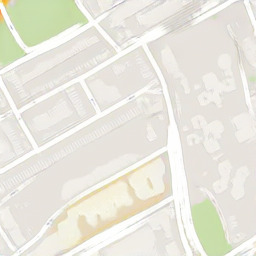}&
\includegraphics[width=.14\textwidth]{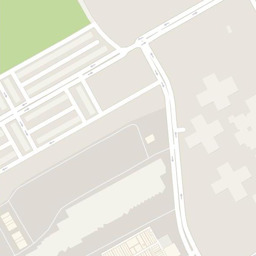} \\ 
  input & Pix2Pix & CycleGAN  & Ours & GT   \\ 
\end{tabular}
}{%
  \captionof{figure}{Unsupervised map prediction visualization.}\label{fig:mapf}
}
\capbtabbox{%
\setlength\tabcolsep{1pt}
  \begin{tabular}{cc} \hline
  Method & Accuracy \\ \hline
  Pix2Pix~\cite{isola2016image} & 43.18\% \\
  CycleGAN~\cite{zhu2017unpaired} & 45.91\% \\
  Ours &  \textbf{46.72\%} \\ \hline
  \end{tabular}
  
}{%
  \caption{Unsupervised map prediction accuracy.}\label{tab:mapt}%
}
\end{floatrow}
\end{figure*} 

\begin{figure*}[!h]
\begin{floatrow}

\capbtabbox{%
\setlength\tabcolsep{1pt}
\begin{tabular}{ccccccc}
\includegraphics[width=.1\textwidth]{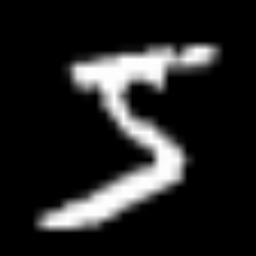}&
\includegraphics[width=.1\textwidth]{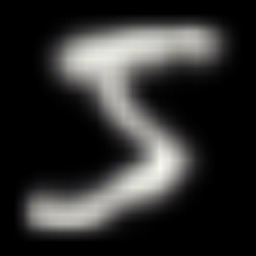}&
\includegraphics[width=.1\textwidth]{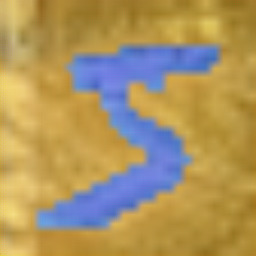}&
\includegraphics[width=.1\textwidth]{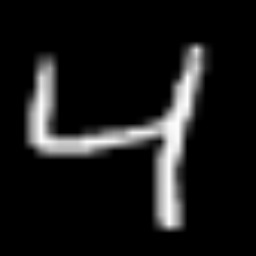}&
\includegraphics[width=.1\textwidth]{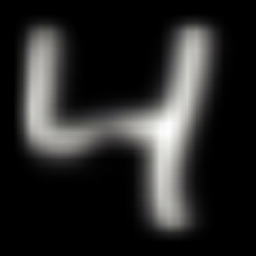}&
\includegraphics[width=.1\textwidth]{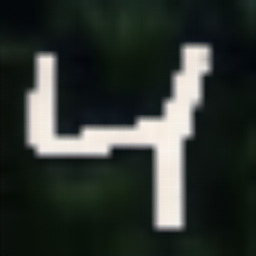} \\
\includegraphics[width=.1\textwidth]{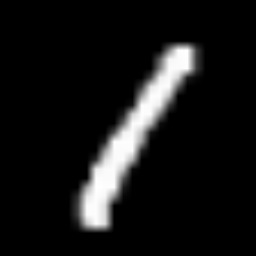}&
\includegraphics[width=.1\textwidth]{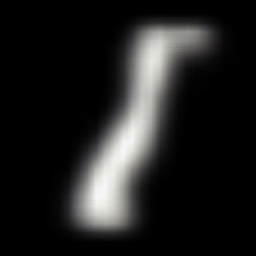}&
\includegraphics[width=.1\textwidth]{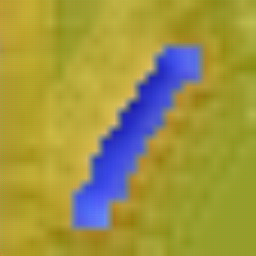}&
\includegraphics[width=.1\textwidth]{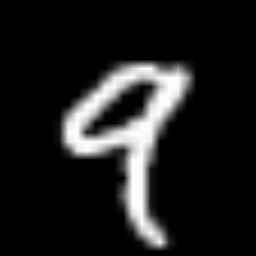}&
\includegraphics[width=.1\textwidth]{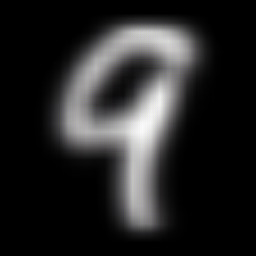}&
\includegraphics[width=.1\textwidth]{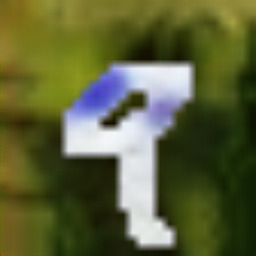} \\
(a) & (b) & (c) & (d) & (e) & (f) \\
\end{tabular}
}{%
  \captionof{figure}{Visualization of image translation from MNIST (a),(d) to USPS (b),(e) and MNIST-M (c),(f).}\label{fig:mnist1}
}
\capbtabbox{%
\setlength\tabcolsep{1pt}
  \begin{tabular}{ccc} \hline
  Method & USPS & MNIST-M \\ \hline
  CoGAN~\cite{liu2016coupled} & 95.65\% & - \\
  PixelDA~\cite{bousmalis2017unsupervised} & 95.90\% & 98.20\% \\
  UNIT~\cite{liu2017unsupervised} & 95.97\% & - \\
  CycleGAN~\cite{zhu2017unpaired} & 94.28\% & 93.16\% \\
  Target-only & 96.50\% & 96.40\% \\
  Ours &  \textbf{96.80\%} & \textbf{98.33\%} \\ \hline
  \end{tabular}
}{%
  \caption{Unsupervised classification results.}\label{fig:mnist2}%
}
\end{floatrow}
\end{figure*} 

\begin{figure*}[!h]
\begin{floatrow}

\capbtabbox{%
\setlength\tabcolsep{1pt}
\begin{tabular}{cccccc}
\includegraphics[width=.12\textwidth]{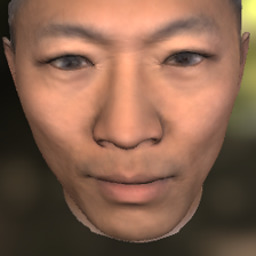}&
\includegraphics[width=.12\textwidth]{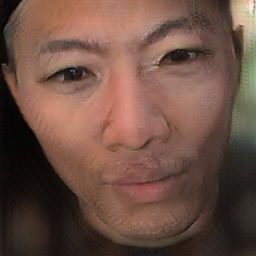}&
\includegraphics[width=.12\textwidth]{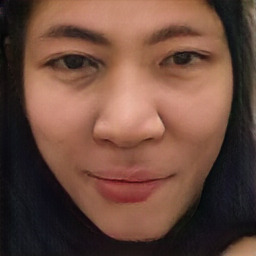}&

\includegraphics[width=.12\textwidth]{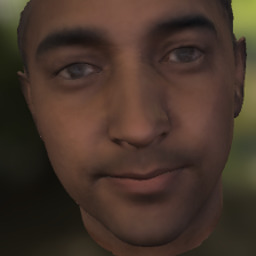}&
\includegraphics[width=.12\textwidth]{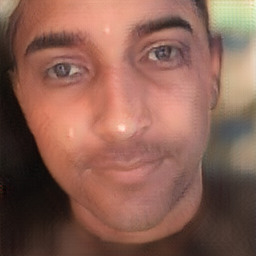}&
\includegraphics[width=.12\textwidth]{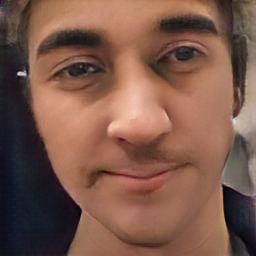}\\
(a) & (b) &(c) & (d) & (e) & (f) \\
 \end{tabular}
}{%
  \captionof{figure}{Visualization of rendered face to real face translation. (a)(d): input rendered faces; (b)(e): CycleGAN results; (c)(f): Our results.}\label{fig:facet2}
}
\capbtabbox{%
\setlength\tabcolsep{1pt}
  \begin{tabular}{cc} \hline
  Method & MSE \\ \hline
  Baseline & 2.26  \\
  CycleGAN~\cite{zhu2017unpaired} & 2.04 \\
  Ours &  \textbf{1.97} \\ \hline
  \end{tabular}
}{%
  \caption{Unsupervised 3DMM prediction results (MSE).}\label{fig:facet}%
}
\end{floatrow}
\end{figure*}

\begin{table}
\begin{center}
\begin{tabular}{cccc} \hline
  { Ours before attn}& {Ours after attn}  &{UNIT}& {CycleGAN} \\ \hline
  {98.90}  &{128.32}  &{241.13}  &{109.36} \\ \hline
\end{tabular}
\caption{FID between generated samples and target domain for horse to zebra.}
\label{tab:fid_hz}
\end{center}
\end{table}
\begin{table}
\begin{center}
\begin{tabular}{cccc} \hline
{   Ours}& {UNIT}&  {CycleGAN} \\ \hline
{  92.86}  &{120.58}  & {102.49} \\ \hline
\end{tabular}
\caption{FID between generated samples and target domain for photo to Van Gogh.}
\label{tab:fid_vg}
\end{center}
\end{table}

\noindent\textbf{Effects of using different layers as feature extractors:} We experimented using different layers of VGG-19 as feature extractors to measure the perceptual loss. Fig.~\ref{fig:ablation} shows visual example of the horse to zebra image translation results trained with different perceptual terms. We can see that only using high-level features as regularization leads to results that are almost identical to the input (Fig.~\ref{fig:ablation} (c)) while only using low-level features as regularization leads to results that are blurry and noisy (Fig.~\ref{fig:ablation} (b)). We find the balance by adopting the first three layers of VGG-19 as feature extractor which does a good job of image translation and also avoids introducing too many noise or artifacts (Fig.~\ref{fig:ablation} (d)).

\subsection{Quantitative Results}

\noindent\textbf{Map prediction:} We translate images from satellite photos to maps with unpaired training data and compute the pixel accuracy of predicted maps. The original photo-map dataset consists of 1096 training pairs and 1098 testing pairs, where each pair contains a satellite photo and the corresponding map. To enable unsupervised learning, we take the 1096 photos from the training set and the 1098 maps from the test set, using them as the training data. Note that no attention is used here since the change is global and we observe training with attention yields similar results. At test time, we translate the test set photos to maps and again compute the accuracy. If the total RGB difference between the color of a pixel on the predicted map and that on the ground truth is larger than 12, we mark the pixel as wrong. Figure~\ref{fig:mapf} and Table~\ref{tab:mapt} show the visual results and the accuracy results, and we can see our approach achieves the highest map prediction accuracy. Note that Pix2Pix is trained with paired data. 

\noindent\textbf{Unsupervised classification: } We show unsupervised classification results on USPS~\cite{denker1989neural} and MNIST-M~\cite{ganin2016domain} in Figure~\ref{fig:mnist1} and Table \ref{fig:mnist2}. On both tasks, we assume we have access to labeled MNIST dataset. We first train a generator that maps MNIST to USPS or MNIST-M and then use the translated image and original label to train the classifier (we do not apply the attention branch here as we did not observe much difference after training with attention). We can see from the results that we achieve the highest accuracy on both tasks, advancing state-of-the-art. The qualitative results clearly show that our MNIST-translated images both preserve the original label and are also visually similar to USPS/MNIST-M. We also notice that our model achieves even better results than the model trained on target labels and conjecture that the classifiers get the benefit of the larger training set size of MNIST dataset.

\noindent\textbf{3DMM face shape prediction: }
As a real-world application of our approach, we study the problem of estimating 3D face shape, which is modeled with the 3D morphable model (3DMM)~\cite{blanz2002face}. 3DMM is widely used for recognition and reconstruction. For a given face, the model encodes its shape with a 100 dimension vector. The goal of 3DMM regression is to predict the 100 dimension vector and we compare them with the ground truth using mean squared error (MSE).~\cite{tran2017regressing} proposes to train a very deep neural network \cite{he2016deep} for 3DMM regression. However, in reality, the labeled training data for real faces are expensive to collect. We propose to use rendered faces instead, as their 3DMM parameters are readily available. We first rendered 200k faces as the source domain and use human selfie photo data of 645 face images we collected as the target domain. For test, we use our collected 112 3D-scanned faces as test data. For the purpose of domain adaptation, we first use our model to translate the rendered faces to real faces and use the results as the training data, assuming the 3DMM parameters stay unchanged. The 3DMM regression model structure is 102-layer Resnet \cite{he2016deep} as in \cite{tran2017regressing}, and was trained with the translated faces. Figure~\ref{fig:facet2} and Table~\ref{fig:facet} show the qualitative results and the final accuracy of 3DMM regression. From the visual results, we see that our translated face preserves the shape of the original rendered face and has higher quality than using CycleGAN. We also reduced the 3DMM regression error compared with baseline (where we trained on rendered faces and tested on real faces) and the CycleGAN results.

\noindent\textbf{Fr\'echet Inception Distance:} We also use the Fr\'echet Inception Distance (FID) \cite{heusel2017gans} between generated samples from our model and target domains for quantitative evaluation. We compute FID for horse to zebra and photo to Van Gogh and results are shown in table \ref{tab:fid_hz} and \ref{tab:fid_vg}. For photo to Van Gogh, we observe that there is no difference between results before and after attention, so we report a single number for our model. The FID results show that our model achieves better FID than baselines for those tasks. For horse to zebra, our model with attention has worse FID than ours without attention and CycleGAN, and we speculate that there might be some correlations between foreground and background in the target domain when computing FID, so using attention might have a negative effect on FID. Also we suspect that FID might not be ideal for image translation task.

\section{Conclusion}
We propose to use a simple model with attention for image translation and domain adaption and achieve superior performance in a variety of tasks demonstrated by both qualitative and quantitative measures. The attention module is particularly helpful to focus the translation on region of interest, remove unwanted changes or artifacts, and may also be used for unsupervised segmentation or saliency detection. Extensive experiments show that our model is both powerful and general, and can be easily applied to solve real-world problems.

%


%

\ifCLASSOPTIONcaptionsoff
  \newpage
\fi



%
\bibliographystyle{ieee}
\bibliography{egbib2}

%
\vspace{-1.4cm}
\begin{IEEEbiography}[{\includegraphics[width=1in,height=1.25in,clip,keepaspectratio]{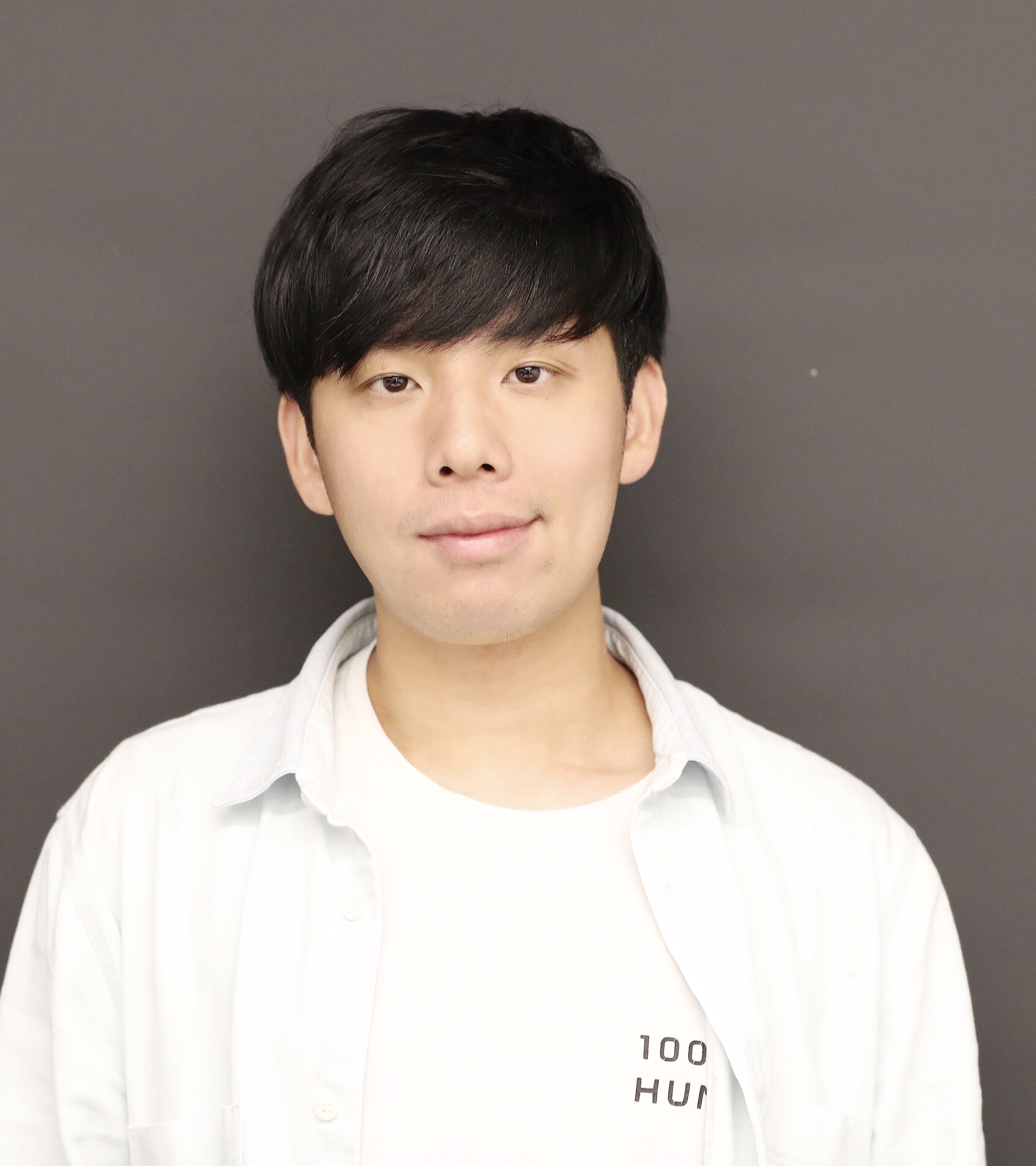}}]{Chao Yang}
is a PhD candidate at the University of Southern California, working with Professor C.-C. Jay Kuo. His current research interests include computer vision and machine learning. He has authored more than 20 conference/journal papers and won the best paper award in ISBA 2018. His work in image inpainting and human avatar digitization has received wide media coverage such as LA Times and Hacker News. He is awarded the prestigious Annenberg PhD Fellowship for his PhD study at USC.
\end{IEEEbiography}
\vspace{-1.2cm}
\begin{IEEEbiography}[{\includegraphics[width=1in,height=1.25in,clip,keepaspectratio]{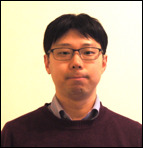}}]{Taehwan Kim}
is a Senior Research Scientist at ObEN Inc. He was a postdoctoral scholar at California Institute of Technology after completing his Ph.D. from Toyota Technological Institute at Chicago, MS from University of Southern California, and BS from Pohang University of Technology (POSTECH). His current research interests are machine learning and applications to computer vision and language.
\end{IEEEbiography}
\vspace{-1.2cm}
\begin{IEEEbiography}[{\includegraphics[width=1in,height=1.25in,clip,keepaspectratio]{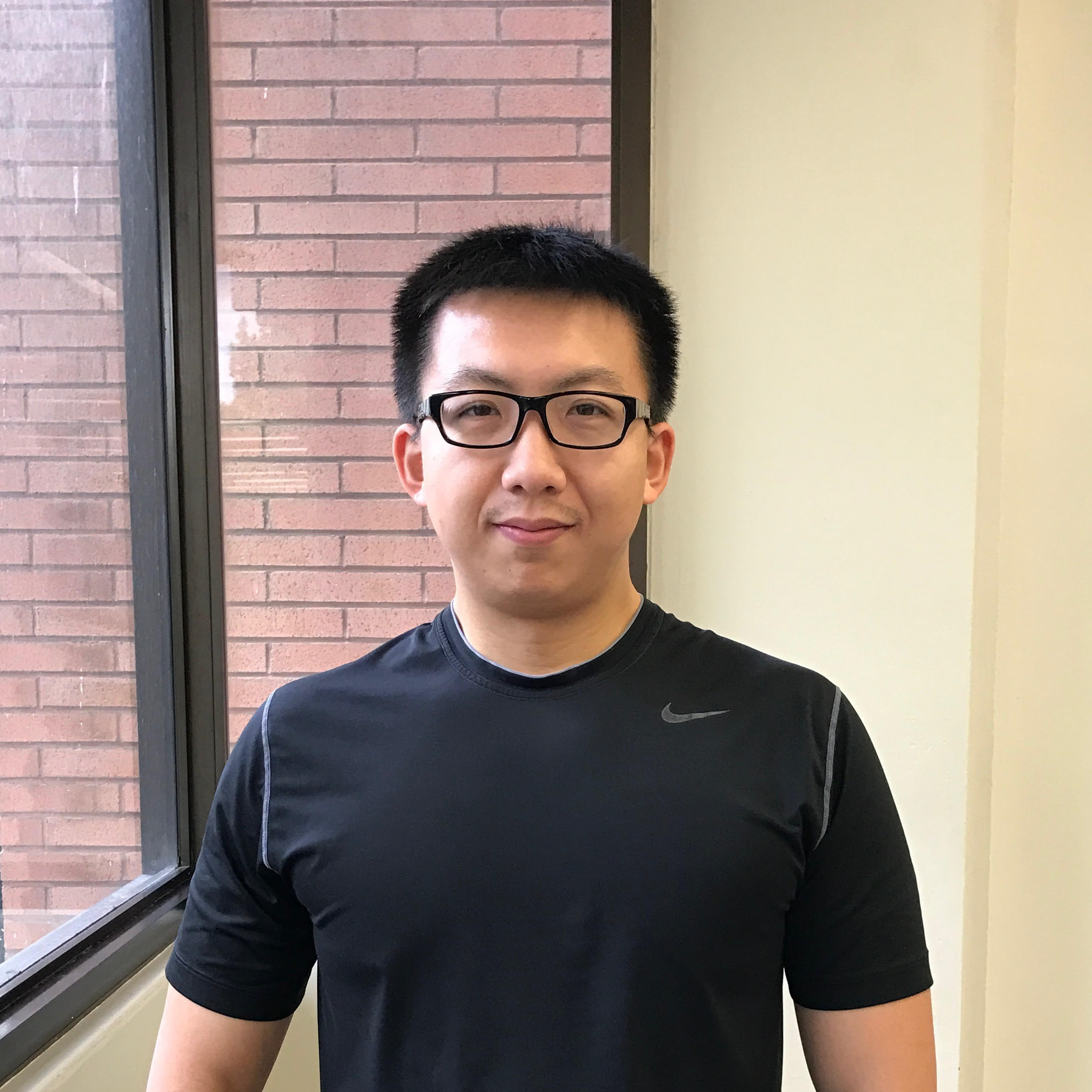}}]{Ruizhe Wang}
Dr. Ruizhe Wang is a Principal Research Scientist at ObEN. Inc, an artificial intelligence company creating intelligent avatars that look, talk and behave like you and is authenticated and secured on the PAI Blockchain. He obtained his Ph.D. from the Computer Science Department at the University of Southern California and received an M.S. from Caltech and B.S. from Tsinghua University. His research interests span computer vision, computer graphics and machine learning. Much of his work has been focusing on developing algorithms to digitize human visual appearance by only using commodity level hardware.
\end{IEEEbiography}
\vspace{-1.2cm}
\begin{IEEEbiography}[{\includegraphics[width=1in,height=1.25in,clip,keepaspectratio]{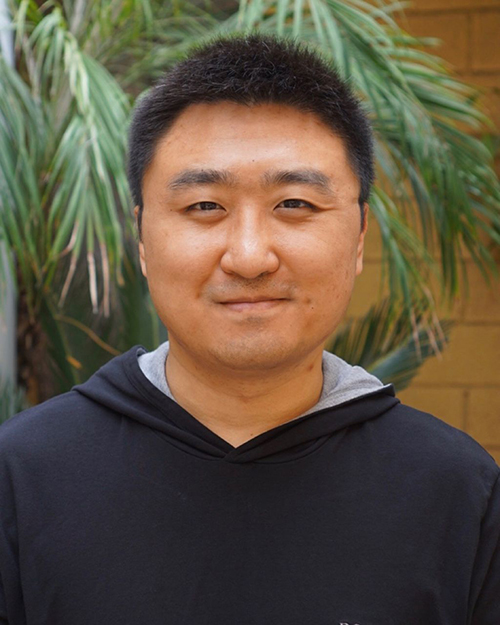}}]{Hao Peng}
Hao Peng is Senior Research Scientist at Oben Inc. His research focus is on computer vision, machine learning and geometric modeling.
\end{IEEEbiography}
\vspace{-1.2cm}

\begin{IEEEbiography}[{\includegraphics[width=1in,height=1.25in,clip,keepaspectratio]{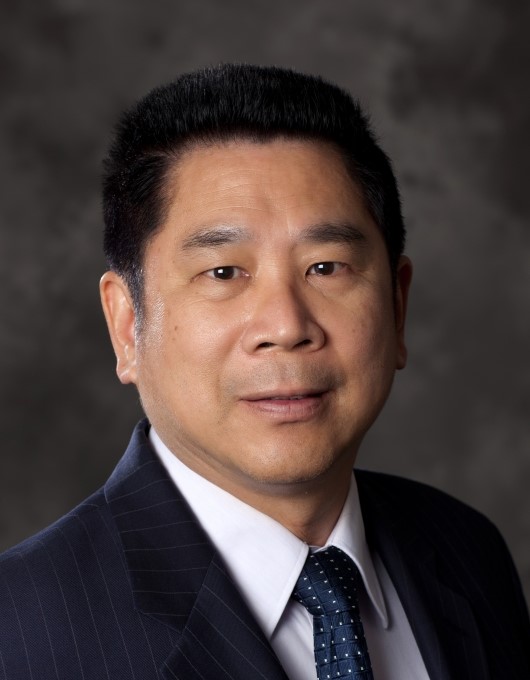}}]{C.-C. Jay Kuo}
C.-C. Jay Kuo (F’99) received the B.S. degree in electrical engineering from National Taiwan University, Taipei, Taiwan, in 1980, and the M.S. and Ph.D. degrees in electrical engineering from the Massachusetts Institute of Technology, Cambridge, in 1985 and 1987, respectively. He is currently the Director of the Multimedia Communications Laboratory and a Professor of Electrical Engineering, Computer Science and Mathematics with the Ming-Hsieh Department of Electrical Engineering, University of Southern California, Los Angeles. He has co-authored about 200 journal papers, 850 conference papers, and ten books. His research interests include digital image/video analysis and modeling, multimedia data compression, communication and networking, and biological signal/image processing. He is a fellow of the American Association for the Advancement of Science and the International Society for Optical Engineers.
\end{IEEEbiography}




\end{document}